\def\year{2020}\relax
\documentclass[letterpaper]{article} 
\usepackage{aaai20}  
\usepackage{times}  
\usepackage{helvet} 
\usepackage{courier}  
\usepackage[hyphens]{url}  
\usepackage{graphicx} 
\usepackage{graphicx}  
\usepackage[utf8]{inputenc} 
\usepackage[T1]{fontenc}    

\usepackage{url}            
\usepackage{booktabs}       
\usepackage{amsmath}
\usepackage{enumerate}
\usepackage{amsfonts}
\usepackage{algorithm}
\usepackage{algorithmic}
\usepackage{nicefrac}       
\usepackage{microtype}      
\usepackage{subfigure}
\usepackage{graphicx}
\usepackage{bbm}
\usepackage{multirow}
\usepackage{xcolor}

\title{Revision in Continuous Space: \\ Unsupervised Text Style Transfer without Adversarial Learning}

\author{
Dayiheng Liu\footnotemark[2], Jie Fu\footnotemark[3], Yidan Zhang\footnotemark[2], Chris Pal\footnotemark[3], Jiancheng Lv\footnotemark[2]\hspace{1mm}\thanks{\hspace{2mm}Correspondence to Jiancheng Lv.} \\
\footnotemark[2]\hspace{0.5mm} College of Computer Science, Sichuan University\\
\footnotemark[3]\hspace{0.5mm} Qu\'ebec Artificial Intelligence Institute (Mila), Polytechnique Montr\'eal\\
{\tt losinuris@gmail.com
 }
 \\
  {\tt lvjiancheng@scu.edu.cn}
  }

\begin{document}
\maketitle
\begin{abstract}
Typical methods for unsupervised text style transfer often rely on two key ingredients: 1) seeking the explicit disentanglement of the content and the attributes, and 2) troublesome adversarial learning. 
In this paper, we show that neither of these components is indispensable. 
We propose a new framework that utilizes the gradients to revise the sentence in a \textit{continuous} space during inference to achieve text style transfer.
Our method consists of three key components: a variational auto-encoder (VAE), some attribute predictors (one for each attribute), and a content predictor.
The VAE and the two types of predictors enable us to perform gradient-based optimization in the \textit{continuous} space, which is mapped from sentences in a \textit{discrete} space, to find the representation of a target sentence with the desired attributes and preserved content. 
Moreover, the proposed method naturally has the ability to simultaneously manipulate multiple \textit{fine-grained} attributes, such as sentence length and the presence of specific words, when performing text style transfer tasks.
Compared with previous adversarial learning based methods, the proposed method is more \textit{interpretable}, \textit{controllable} and \textit{easier} to train.
Extensive experimental studies on three popular text style transfer tasks show that the proposed method significantly outperforms five state-of-the-art methods.
\end{abstract}

\section{Introduction}
Text style transfer, which is an under-explored challenging task in the field of text generation, aims to convert some attributes of a sentence (e.g., negative sentiment) to other attributes (e.g., positive sentiment) while preserving attribute-independent content. 
In other words, text style transfer can generate sentences with desired attributes in a controlled manner. Due to the difficulty in obtaining training sentence pairs with the same content and differing styles, this task usually works in an unsupervised manner where the model can only access non-parallel, but style labeled sentences.

Most existing methods \cite{hu2017toward,shen2017style,fu2018style,li2018delete} for text style transfer usually first explicitly disentangle the content and the attribute through an adversarial learning paradigm \cite{Goodfellow2014Generative}. The attribute-independent content and the desired attribute vector are then fed into the decoder to generate the target sentence. However, some recent evidence suggests that using adversarial learning may not be able to learn representations that are disentangled \cite{li2018delete,lample2019rewriting}. Moreover, vanilla adversarial learning is designed for generating real-valued and continuous data but has difficulties in generating sequences of discrete tokens directly . As a result, algorithms such as REINFORCE \cite{Sutton1999Policy} or those that approximate the discrete tokens with temperature-softmax probability vectors \cite{kusner2016gans,Zhang2017Adversarial} are used. Unfortunately, these methods tend to be unstable, slow, and hard-to-tune in practice \cite{lample2019rewriting}. 

Is it really a necessity to explicitly disentangle the content and the attributes? Also, do we have to use adversarial learning to achieve text style transfer?
Recently, the idea of mapping the discrete input into a continuous space and then performing gradient-based optimization with a predictor to find the representation of a new discrete output with desired property has been applied for sentence revision \cite{mueller2017sequence} and neural architecture search \cite{luo2018neural}. Motivated by the success of these works, we propose a new solution to the task of content-preserving text style transfer. 

The proposed approach contains three key components: (a) A \textbf{variational auto-encoder (VAE)} \cite{kingma2013auto}, whose encoder maps sentences into a smooth continuous space and its decoder can map a continuous representation back to the sentence. (b) Some \textbf{attribute predictors} that take the continuous representation of a sentence as input and predict the attributes of its decoder output sentence, respectively. These attribute predictors enable us to find the target sentence with the desired attributes in the continuous space. (c) A \textbf{content predictor} that takes the continuous representation of a sentence as input and predicts the Bag-of-Word (BoW) feature of its decoder output sentence. The purpose of component (c) is threefold: First, it could enhance the content preservation during style transfer; Second, it enables the target sentence to contain some specific words; Third, it can tackle the vanishing latent variable problem of VAE \cite{zhao2017learning}. With the gradients obtained from these predictors, we can revise the continuous representation of the original sentence by gradient-based optimization to find a target sentence with the desired fine-grained attributes, and achieve the content-preserving text style transfer.

The method we propose has three major advantages compared to previous methods:
\begin{itemize}
\item The method can be \textit{easily} trained on the non-parallel dataset, avoiding the problem of training difficulties caused by adversarial learning and achieving higher performance. 
\item  Unlike previous methods directly generate the target-style sentence through once feed-forward in the inference stage, our method revises the original sentence with gradient information for several steps during inference, which explicitly presents the process of the style transfer and can easily provide us multiple results with tuning the gradients.
Therefore, the proposed method has higher \textit{interpretability} and is more \textit{controllable}.
\item Most previous text style transfer methods that only control a single binary attribute (e.g., positive and negative sentiments).  In contrast, our approach is more \textit{generic} in the sense that it naturally has the ability to control multiple fine-grained attributes, such as sentence length and the existence of specific words. 
\end{itemize}

Extensive experimental comparisons on three popular text style transfer tasks show that the proposed method significantly outperforms five state-of-the-art methods. The source code is available at \url{https://github.com/dayihengliu/Fine-Grained-Style-Transfer}.

\begin{figure*}
\centering
\includegraphics[width=0.90\linewidth]{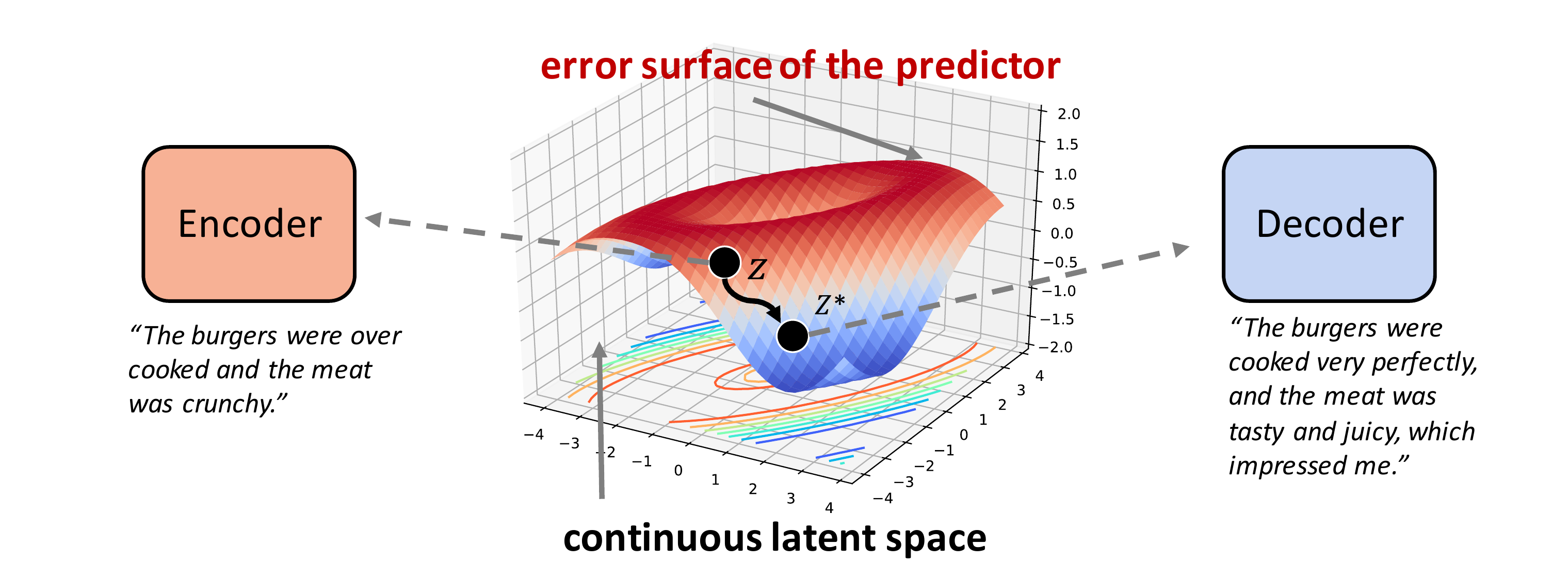}
\caption{There is an example of content-preserving text sentiment transfer, and we hope to further increase the length of the target sentence compared with the original sentence. The original sentence $x$ with negative sentiment is mapped to continuous representation $z$ via encoder. Then $z$ is revised into $z^*$ by minimizing the error $\mathcal{L}_{\text{Attr,$s_1$}}(\theta_{s_1};s_1=\{\text{sentiment}=\textit{positive}\})+\mathcal{L}_{\text{Attr,$s_2$}}(\theta_{s_2};s_2=\{\text{length}=\textit{20}\})+\lambda_\text{bow}\mathcal{L}_{\text{BOW}}(\theta_\text{bow};x_\text{bow}=[\textit{burgers}, \textit{meat}])$ with the sentiment predictor $f_1$, length predictor $f_2$, and the content predictor $f_\text{bow}$. Afterwards the target sentence $x^*$ is generated by decoding $z^*$ with beam search via decoder [best viewed in color].}
\label{fig:frame}
\end{figure*}

\section{Methodology}
Let $\mathcal{D}=\{(x^1, s^1),...,(x^n, s^n)\}$ denotes a dataset which contains $n$ sentences $x^i$ paired with a set of attributes $s^i$. Each $s$ has $k$ attributes of interest $s=\{s_1,...,s_k\}$. Unlike most previous methods \cite{shen2017style,fu2018style,prabhumoye2018style,li2018delete,yang2018unsupervised} that only consider a single binary attribute (e.g., positive or negative sentiments), our approach naturally has the ability to control multiple fine-grained attributes during style transfer.
Here we take two fine-grained attributes, sentence length and the presence of specific words (e.g., a pre-defined subject noun), as the case study.
For example, given a original sentence $x$ =``\textit{the salads are fresh and delicious.}'', its attribute set can be $s$=\{sentiment=\textit{positive}, length=\textit{7}, subject\_noun=\textit{salads}\}. Our task is to learn a generative model $G$ that can generate a new sentence $x^*$ with the required attributes $s^*$, and retain the attribute-independent content of $x$ as much as possible.

\subsection{Model Structure}
The proposed model consists of three components: a variational auto-encoder (VAE), attribute predictors, and a content predictor.

\noindent \textbf{Variational auto-encoder} $G$. The VAE integrates stochastic latent representation $z$ into the auto-encoder architecture. Its RNN encoder maps a sentence $x$ into a continuous latent representation $z$:
\begin{equation}
z\sim G_\text{enc}(\theta_\text{enc};x)=q_E(z|x),
\end{equation}
and its RNN decoder maps the representation back to reconstruct the sentence $x$:
\begin{equation}
x\sim G_\text{dec}(\theta_\text{dec};z)=p_G(x|z),
\end{equation}
where $\theta_\text{enc}$ and $\theta_\text{dec}$ denote the parameters of the encoder and decoder. The VAE is then optimized to minimize the reconstruction error $\mathcal{L}_\text{rec}$ of input sentences, and meanwhile minimize the KL term $\mathcal{L}_\text{KL}$ to encourages the $q_E(z|x)$ to match the prior $p(z)$:
\begin{equation}
\begin{split}
&\mathcal{L}_{\text{VAE}}(\theta_\text{enc}, \theta_\text{dec}) = \mathcal{L}_\text{rec} + \mathcal{L}_\text{KL} \\ &= -\mathbb{E}_{q_E(z|x)}\left[\log p_{G}(x|z) \right] +\mathcal{D}_\text{KL}(q_{E}(z|x) \| p(z)) , 
\end{split} 
\end{equation}
where $\mathcal{D}_\text{KL}(\cdot \| \cdot)$ is the KL-divergence. Compared with traditional deterministic auto-encoder, the VAE offers two main advantages in our approach: 

(1) Deterministic auto-encoders often have ``holes'' in their latent space, where the latent representations may not able to generate anything realistic \cite{RobertsERHE18}. In contrast, by imposing a prior standardized normal distribution $\mathcal{N}(z; 0, I)$ on the latent representations, the VAE learns latent representations not as single isolated points, but as soft dense regions in continuous latent space which makes it be able to generate plausible examples from every point in the latent space \cite{bowman2015generating}. This characteristic avoids the problem that the representation $z^*$ revised (optimized) by the gradient not being able to generate a plausible sentence. 

(2) This continuous and smooth latent space learned by the VAE enables the sentences generated by adjacent latent representation to be similar in content and semantics \cite{bowman2015generating,semeniuta2017hybrid,goyal2017z,yang2017improved,shen2018improving}. Therefore, if we revise the representation $z$ within a reasonable range (i.e., small enough), the resulting new sentence would not differ much in content from the original sentence.

\noindent \textbf{Attribute predictors} $f_1,...,f_k$. Each of them takes the representation $z$ as input and predict one attribute $s_j$ of the decoder output sentence $\hat{x}$ generated by $z$. For example, the attribute predictor can be a binary classifier for positive-negative sentiment prediction or a regression model for sentence length prediction. With the gradients provided by the predictors, we can revise the continuous representation $z$ of the original sentence $x$ by gradient-based optimization to find a target sentence $x^*$ with the desired attributes $s^*$.

The attribute predictors $f_1,...,f_k$ are jointly trained with VAE. For M-classification predictors, we have
\begin{equation}
\begin{split}
\mathcal{L}_{\text{Attr}, s_j}(\theta_{s_j}, \theta_\text{enc}) =  - \mathbb{E}_{q_E(z|x)} \log \left[ f_j(z) \right],
\end{split}
\end{equation}
where $f_j(z)= \text{MLP}_j(z) = p(s_j|z) \in \mathbb{R}^\text{M}$. And for the regression predictors, we have
\begin{equation} 
\mathcal{L}_{\text{Attr}, s_j}(\theta_{s_j}, \theta_\text{enc}) = \mathbb{E}_{q_E(z|x)} \left[ (s_j - f_j(z))^2 \right], 
\end{equation}
where $f_j(z) = \text{MLP}_j(z) \in \mathbb{R}^1$. In this joint training, we take the attributes of the input sentence $x$ as the label of predictors. Since the predictor are designed to predict the attribute of the sentence $\hat{x}$ generated by $z$, we further train each predictor individually after joint training. We sample $z$ from $\mathcal{N}(z; 0, I)$ and feed it into the decoder to generate a new sentence $\hat{x}$. Afterwards we feed $\hat{x}$ into the CNN text classifiers \cite{kim2014convolutional} which are trained on the training set to predict its attributes\footnote{Some attributes can be obtained directly without using classifiers, such as the length $\hat{s}_j$ of $\hat{x}$.} as the label of the predictors:
\begin{equation} 
\begin{split}
& \mathcal{L}'_{\text{Attr}, s_j}(\theta_{s_j}) =  - \mathbb{E}_{p(z)p_G(\hat{x}|z)} \log \left[ p(\text{CNN}(\hat{x})|z) \right], \\
& \mathcal{L}'_{\text{Attr}, s_j}(\theta_{s_j}) = \mathbb{E}_{p(z)p_G(\hat{x}|z)} \left[ (\hat{s}_j - f_j(z))^2 \right]. 
\end{split}
\end{equation}

\noindent \textbf{Content predictor} $f_{\text{bow}}$. It is a multi-label classifier that takes $z$ as input and predicts the Bag-of-Word feature $x_\text{bow}$ of its decoder output sentence:
\begin{equation} 
f_{\text{bow}}(z) = \text{MLP}_\text{bow}(z) = p(x_{\text{bow}}|z). 
\end{equation}
We assume $p(x_{\text{bow}}|z)$ as $|x|$-trial multimodal distribution:
\begin{equation}
\begin{split}
 \log p(x_\text{bow}|z) = \log \prod_{t=1}^{|x|}\frac{e^{f_\text{bow}^{(x_t)}}}{\sum_j^\mathcal{V} e^{f_\text{bow}^{(x_j)}}}, \\
\end{split}
\end{equation}
where $\mathcal{V}$ is the size of vocabulary, $|x|$ is the length of $x$, and $f_\text{bow}^{(x_j)}$ is the output value of $j$-th word in $f_\text{bow}\in \mathbb{R}^\mathcal{V}$. 

The training of content predictor $f_{\text{bow}}$ is similar to attribute predictors. It is jointly trained with VAE:
\begin{equation} 
\begin{split}
\mathcal{L}_{\text{BOW}}(\theta_\text{bow},\theta_\text{enc}) =  - \mathbb{E}_{q_E(z|x)} \log \left[ p(x_\text{bow}|z) \right].
\end{split}
\end{equation}
After joint training, it is trained separately through:
\begin{equation} 
\begin{split}
\mathcal{L}'_{\text{BOW}}(\theta_\text{bow}) =  - \mathbb{E}_{p(z)p_G(\hat{x}|z)} \log \left[ p(\hat{x}_\text{bow}|z) \right].
\end{split}
\end{equation}

During text style transfer, we can similarly revise the representation $z$ with the gradient provided by the content predictor $f_{\text{bow}}$ to enhance content preservation. Here we consider two ways to enhance content preservation during style transfer. We can set $x_\text{bow}$ to contain all the words in the original sentence $x$, which means that we try to find a sentence $x^*$ with the desired attributes $s^*$ and keep all the words of the original sentence as much as possible to achieve content preservation. However, retaining all the words is often not what we want. For example, $x^*$ should not contain the original emotional words in the task of text sentiment transfer. Instead, the noun in the original sentence should be retained in such a task \cite{melnyk2017improved,li2018delete,john2019disentangled}. Therefore, we can set $x_\text{bow}$ to contain only all nouns in $x$. Furthermore, we can set $x_\text{bow}$ to contain some desired specific words to achieve finer-grained control of target sentences.

\noindent \textbf{Putting them together}, the final joint training loss $\mathcal{L}$ is as follows:
\begin{equation} 
\mathcal{L} = \mathcal{L}_\text{VAE} + \lambda_b \mathcal{L}_{\text{BOW}} + \lambda_{s} \sum_{j=1}^k{\mathcal{L}_{\text{Attr}, s_j}},
\end{equation}
where $\lambda_b$ and $\lambda_{s}$ are balancing hyper-parameters. It should be noted that $\mathcal{L}_{\text{BOW}}$ and $\mathcal{L}_{\text{Attr}, s_j}$ also act as regularizers that prevent the encoder from being trapped into a KL vanishing state \cite{bowman2015generating,kingma2016improved,yang2017improved,shen2018improving,alemi2018fixing}.

\subsection{Text Style Transfer}
Given the original sentence $x$, the inference process of style transfer is performed in the continuous space. We revise its representation $z$ by gradient-based optimization as follows:

\begin{equation} 
\hat{z} = z - \eta (\sum_{j=1}^k{\nabla_z \mathcal{L}_{\text{Attr}, s_j}} + \lambda_c \nabla_z \mathcal{L}_{\text{BOW}}),
\end{equation} 
where $\eta$ is the step size and $\lambda_c$ is the trade-off parameter to balance the content preservation and style transfer strength. We iterate such optimization to find the $z^*$ until the output confidence score of attribute predictors $p(s_j |z)$ is greater than a threshold $\beta$ or reach the maximum number of rounds $T$. The target $x^*$ is obtained by decoding $z^*$ with a beam search \cite{och2004alignment}. An example procedure is shown in Figure \ref{fig:frame}.

\begin{table*}[t]\small
\begin{center}
\begin{tabular}{lcccccccccccc}
\hline
Methods & Accuracy$\uparrow$ & PPL$\downarrow$ & Overlap$\uparrow$ & Noun\%$\uparrow$ & BLEU$\uparrow$ \\
 \hline
Original & \underline{0.1} & 22.9 & 100.0 & 100.0 & 42.4 \\
Human & 91.8 & 76.9 & 47.2 & 78.5 & 100.0 \\
\hline
Delete, Retrieve, \& Generate \cite{li2018delete}: & \\
TemplateBased & 81.3 & \underline{183.6} & 55.6 & \textbf{83.3} & 28.9 \\
DeleteOnly & 85.8 & \underline{81.4} & 49.5 & 74.9 & 24.7 \\
DeleteAndRetrieve & 89.5 & \underline{96.1} & 49.4 & 74.0 & 24.9  \\
RetrievalOnly & \textbf{98.4} & 25.7 & \underline{15.8} & \underline{39.6} & \underline{4.7} \\
\hline
StyleEmbedding \cite{fu2018style} & \underline{7.2} & \underline{93.9} & \textbf{75.4} & 74.2 & \textbf{31.9} \\
MultiDecoder \cite{fu2018style} & \underline{48.8} & \underline{166.5} & 51.5 & 52.2 & 23.1 \\
BTS \cite{prabhumoye2018style} & 94.8 & 32.8 & \underline{21.5} & \underline{23.5} & \underline{6.8} \\
CrossAligned \cite{shen2017style} & \underline{73.6} & \underline{72.0} & 41.1 & \underline{42.9} & 18.4 \\
\hline
Ours (content-strengthen) & 88.2 & 26.5 & 46.6 & 77.4 & 21.8  \\ 
Ours (style-content balance) & 92.3 & \textbf{18.3} & 38.9 & 69.3 & 18.8 \\ 
Ours (style-strengthen) & 95.7 & 20.6 & 39.7 & 61.5 & 17.9 \\ \hline
\hline
Methods & Accuracy$\uparrow$ & PPL$\downarrow$ & Overlap$\uparrow$ & Noun\%$\uparrow$ & BLEU$\uparrow$ \\
 \hline
Original & \underline{23.4} & 24.4 & 100.0 & 100.0 & 57.2  \\
Human & 88.1 & 62.9 & 60.5 & 85.0 & 100.0  \\
\hline
Delete, Retrieve, \& Generate \cite{li2018delete}: & \\
TemplateBased & \underline{69.6} & \underline{108.9} & 73.3 & 87.9 & 42.8  \\
DeleteOnly & \underline{51.6} & 49.3 & \textbf{74.4} & \textbf{95.1} & \textbf{44.7}  \\
DeleteAndRetrieve & \underline{55.2} & 48.2 & 69.1 & 92.6 & 41.8  \\
RetrievalOnly & 87.2 & 28.7 & \underline{21.0} & 44.5 & \underline{6.7} \\
\hline
StyleEmbedding \cite{fu2018style} & \underline{40.5} & \underline{87.7} & 42.2 & 41.8 & 22.1  \\
MultiDecoder \cite{fu2018style} & \underline{66.5} & \underline{80.8} & 30.6 & 30.4 & 14.3  \\
BTS \cite{prabhumoye2018style} & 82.6 & 25.3 & \underline{24.7} & \underline{22.5} &
\underline{9.2}  \\
CrossAligned \cite{shen2017style} & \underline{69.6} & 18.3 & \underline{19.3} & \underline{20.4} & \underline{5.0}  \\
\hline
Ours (content-strengthen) & 81.9 & 35.0 & 37.7 & 76.0 & \underline{11.5} \\
Ours (style-content balance) & 85.1 & 21.8 & 49.3 & 49.8 & 21.5  \\ 
Ours (style-strengthen) & \textbf{90.0} & \textbf{15.9} & 39.5 & 41.4 & 16.3  \\ 
\hline
 \end{tabular}
\end{center}
\caption{\label{tab:sentiment} Evaluation results of the sentiment transfer tasks on Yelp (Top) and Amazon (Bottom). The notation $\uparrow$ means the higher the better, while $\downarrow$ means the lower the better. For our models, we report different results (denoted as Ours (content-strengthen), Ours (style-content balance), and Ours (style-strengthen)) corresponding to different choices of hyper-parameters ($\lambda_\text{c}$ and $\beta$), which demonstrates our models' ability to control the trade-off between attribute transfer and content preservation. For each evaluation criterion, we bold the best values (except for Human and Original). The accuracies of the classifier on the test set of Yelp and Amazon are 98.2\% and 84.0\%. Note that a good model should perform well on all metrics, we further highlight the metrics where the performances of the models are poor with underline.} 
\end{table*}

\section{Experiments}
In this section, we evaluate the proposed method on three publicly available datasets of sentiment transfer and gender style transfer tasks. Then we conduct several experiments on text sentiment transfer tasks and simultaneously control other fine-grained attributes such as length and keyword presence.

\subsection{Text Sentiment Transfer} \label{sec:tst}
\paragraph{Data} We use two datasets, Yelp restaurant reviews and Amazon product reviews \cite{he2016ups}\footnote{These datasets can be download at \url{http://bit.ly/2LHMUsl}.}, which are commonly used in prior works too \cite{shen2017style,fu2018style,li2018delete,prabhumoye2018style}. Following their experimental settings, we use the same pre-processing steps and similar experimental configurations.

\paragraph{Metrics} There are three criteria for a good style transfer \cite{li2018delete,prabhumoye2018style}. Concretely, the generated sentences should: 1) have the desired attributes; 2) be fluent; 3) preserve the attribute-independent content of the original sentence as much as possible. For the first and second criteria, we follow previous works \cite{shen2017style,fu2018style,li2018delete,prabhumoye2018style} in using model-based evaluation. We measure whether the style is successfully transferred according to the prediction of a pre-trained bidirectional LSTM classifier \cite{schuster1997bidirectional}, and measure the language quality by the perplexity (PPL) of the generated sentences with a pre-trained language model. Following previous works, we use the trigram Kneser-Ney smoothed language model \cite{kneser1995improved} trained on the respective dataset. Since it is hard to measure the content preservation, we follow previous works and report two metrics: 1) Word overlap, which counts the unigram word overlap rate of the original sentence $x$ and the generated sentence $\hat{x}$, computed by $\frac{\text{count}(w_x\cap w_{\hat{x}})}{\text{count}(w_x\cup w_{\hat{x}})}$; 2) As argued in \cite{melnyk2017improved,li2018delete}, almost all of the nouns in sentences are attribute-independent content and should be kept in style transfer task, we also calculate the percentage of nouns (e.g., as detected by a POS tagger) in the original sentence appearing in the generated sentence (denoted as Noun\%). 
There are 1000 human annotated sentences as the ground truth of the transferred sentences in \cite{li2018delete}. We also take them as references and report the bi-gram BLEU scores \cite{papineni2002bleu}.

\paragraph{Baselines} We compare our method with several previous state-of-the-art methods \cite{shen2017style,fu2018style,li2018delete,prabhumoye2018style}. We report the results of the human-written sentences as a strong baseline. The results of not making any changes to the original sentences (denoted as Original) are also reported. The effect of using different hyper-parameters and the ablation study of our method are analyzed in \textbf{Appendix A}.
\paragraph{Results} Table \ref{tab:sentiment} shows the evaluation results on two datasets. 
It should be noted that a good style transfer method should perform well on all metrics as we discussed above.
If we only use the original sentence as the output without any modifications, we can still get good performance on both language fluency (PPL) and content retention (Overlap, Noun\%) as shown in the first row of Table \ref{tab:sentiment}.
Therefore, we highlight the metrics where the performances of the models are poor with underline.
We find that StyleEmbedding and MultiDecoder achieve high content retention (Overlap, BLEU, and Noun\%), but their fluency (PPL) and transfer accuracy are significantly worse than our method. 
Though the fluency of CrossAligned is better than StyleEmbedding and MultiDecoder, it does not perform well in both content preservation and sentiment transfer.
On the contrary, BST achieves high fluency and transfer accuracy, while the content is poorly preserved. 
Ours (style-strengthen) performs better than BST and CrossAligned on all metrics for these two tasks. 

Because the methods proposed in \cite{li2018delete} (except for RetrievalOnly) are based on prior knowledge, which directly revises few words in the original sentence in the discrete space, they can easily achieve high content retention but do not guarantee fluency and accuracy. As shown in the results, their fluency and the transfer accuracy are bad compared to our method. 
The method RetrievalOnly retrieves the human-written sentence as output, thus this method can achieve high transfer accuracy and fluency, but its content retention is worse than our method.
Our methods revise the original sentence in a continuous space, which does well in fluency, content preservation, and transfer accuracy. In addition, our methods can control the trade-off between the transfer accuracy and content preservation.

\begin{table*}[ht] \small
\begin{center}
\begin{tabular}{@{~~~~~}l@{~~~~~}l@{~~~~~}}
\toprule
\multicolumn{2}{c}{Sentiment transfer from \textbf{negative} to \textbf{positive} (Yelp)}\\
\toprule
Original &  we sit down and we got some really slow and lazy service . \\
Human & the service was quick and responsive . \\
CrossAligned & we went down and we were a good , friendly food . \\
MultiDecoder & we sit down and we got some really and fast food . \\
DeleteAndRetrieve & we got very nice place to sit down and we got some service . \\
BackTranslation & we got and i and it is very nice and friendly staff . \\
Ours (content-strengthen) & we sat down and got some really good service and friendly people . \\
Ours (style-content balance) & we sat down the street and had some really nice and fast service . \\
Ours (style-strengthen) & we really sit down and the service and food were great . \\
\toprule
\multicolumn{2}{c}{Sentiment transfer from \textbf{positive} to \textbf{negative} (Yelp)}\\
\toprule
Original &  i love this place , the service is always great ! \\
Human & hate this place , service was bad . \\
CrossAligned & i know this place , the food is just a horrible ! \\
MultiDecoder & i love this place , the service is always great ! \\
DeleteAndRetrieve & i did not like the homework of lasagna , not like it , . \\
BackTranslation & i wish i have been back , this place is a empty ! \\
Ours (content-strengthen) & however , this place is the worst i have ever been to . \\
Ours (style-content balance) & i do n't know why i love this place , but the service is horrible . \\
Ours (style-strengthen) & i do n't know why this place has the worst customer service ever . \\
\toprule
\end{tabular}
\end{center}
\caption{Samples of the sentiment transfer task from ours and baselines on Yelp. The Original denotes the input sentence, and the Human denotes the human annotated sentence. The samples of the sentiment transfer from negative to positive and positive to negative are shown in top and bottom, respectively.}\label{tab:sentiment_yelp}
\end{table*}

\paragraph{Human Evaluation}
We conduct human evaluations to further verify the performance of our methods on two datasets further. Following previous works \cite{li2018delete,fu2018style}, we randomly select 50 original sentences and ask 7 evaluators\footnote{All evaluators have Bachelor or higher degree. They are independent of the authors' research group.} to evaluate the sentences generated by different methods. Each generated sentence is rated on the scale of 1 to 5 in terms of transfer accuracy, preservation of content, and language fluency. The results are shown in Table \ref{tab:human}. It can be seen that our models perform well on all metrics and significantly outperform all baselines on the percentage success rate (Suc\%) for two datasets. The generated examples can be found in \textbf{Appendix B}.

\begin{table*}[h] \small
\centering
\begin{tabular}{|l|c|c|c|c|c|c|c|c|c|c|c|c}
\hline
\multirow{2}{*}{} & \multicolumn{4}{c|}{Yelp} & \multicolumn{4}{c|}{Amazon} \\ \cline{2-9} 
& Acc  & Gra  & Con  & Suc\%  & Acc  & Gra  & Con  & Suc\%  \\ \hline
Human & 4.1 & 4.4 & 3.6 & 78 & 3.5 & 4.3 & 3.9 & 60 \\ \hline
CrossAligned \cite{shen2017style} & 3.3 & 2.9 & 2.6 & 22 & 3.0 & 3.3 & 1.6 & 6 \\
MultiDecoder \cite{fu2018style} & 2.4 & 3.0 & 3.1 & 12 & 2.3 & 2.7 & 2.5 & 6 \\
BTS \cite{prabhumoye2018style} & \textbf{3.9} & 3.7 & 1.8 & 26 & 2.8 & 3.3 & 1.8 & 8 \\
DeleteAndRetrieve \cite{li2018delete} & 3.8 & 3.6 & \textbf{3.5} & 54 & 2.4 & 3.5 & \textbf{3.8} & 28 \\
\hline
Ours (content-strengthen) & 3.6 & 4.1 & 3.1 & 66 & 3.4 & 4.0 & 2.8 & 42 \\
Ours (style-content balance) & 3.7 & \textbf{4.3} & 3.2 & \textbf{72} & 3.7 & 4.0 & 2.4 & 40 \\
Ours (style-strengthen) & 3.8 & 4.1 & 3.0 & 60 & \textbf{3.8} & \textbf{4.5} & 2.5 & \textbf{50} \\
\hline
\end{tabular} 
\caption{\label{tab:human} Human evaluation results of the sentiment transfer tasks on Yelp and Amazon. We show average human ratings for transfer accuracy (Acc), preservation of content (Con), and fluency of sentences (Gra) on 1 to 5 score. ``Suc\%" denotes the overall percentage success rate. Similar to previous works, we consider a generated output ``successful" if it is rated no less than 3 on all three criteria (Att, Con, Gra).}
\end{table*}

\begin{table*}[!ht]
\small
\centering
\begin{tabular}{lcccccccccccc}
\hline
Methods & Accuracy$\uparrow$ & PPL$\downarrow$ & Overlap$\uparrow$ & Noun\%$\uparrow$ \\
 \hline
Orginal & 21.9 & 183.4 & 100.0 & 100.0  \\
BTS \cite{prabhumoye2018style} & 60.3 & 145.0 & 37.9 & 35.3  \\
Ours (content-strengthen) & 70.6 & 98.2 & 46.8 & \textbf{69.6}  \\
Ours (style-content balance) & 71.3 & 87.8 & \textbf{51.8} & 57.5  \\
Ours (style-strengthen) & \textbf{79.9} & \textbf{78.9} & 46.4 & 53.8\\
\hline
 \end{tabular}
\caption{\label{tab:gender} Evaluation results of the gender transfer task on Yelp. For our models, we report different results corresponding to different choices of hyper-parameters ($\lambda_\text{c}$ and $\beta$) to demonstrate our models' ability to control the trade-off between attribute transfer and content preservation. The accuracy of the classifier on the test set is 83.1\%.}
\end{table*}

\begin{table*}[h] \small
\begin{center}
\begin{tabular}{lcccccccccccc}
\hline
Methods & Accuracy$\uparrow$ & PPL$\downarrow$ & Overlap$\uparrow$ & Noun\%$\uparrow$ & Len\% & Key\%$\uparrow$ \\
 \hline
Original & 0.1 & 22.9 & 100.0 & 100.0 & 100.0 & 7.8 \\ 
\hline
Keywords &  16.7 & 43.9 & \textbf{39.2} & 56.0 & 98.1 & \textbf{92.3} \\
Sentiment + Keywords  & 91.6 & 52.6 & 24.5 & 42.4 & 106.0 & 78.3\\ \hline
Length$\Uparrow$ & 0.2 & 29.8 & 25.0 & 48.3 & \textbf{208.8} & 5.9\\ 
Sentiment + Length$\Uparrow$ & \textbf{97.7} & 25.4 & 21.4 & 51.7 & 189.5 & 9.2\\ 
Keywords + Length$\Uparrow$ & 25.6 & 44.5 & 29.8 & \textbf{61.8} & 165.0 & 83.2\\ 
Sentiment + Keywords + Length$\Uparrow$  & 93.0 & 51.8 & 18.8 & 50.0 & 183.7 & 66.6\\ \hline 
Length$\Downarrow$ & 0.2 & 31.3 & 30.7 & 25.2 & \textbf{40.8} & 6.3\\ 
Sentiment + Length$\Downarrow$  & 95.1 & \textbf{23.0} & 29.1 & 38.1 & 66.9 & 6.7\\ 
Keywords + Length$\Downarrow$ & 21.4 & 87.0 & 28.4 & 38.9 & 61.6 & 83.7\\ 
Sentiment + Keywords + Length$\Downarrow$ & 87.6 & 123.8 & 16.3 & 23.7 & 60.9 & 63.0\\
\hline
 \end{tabular}
\end{center}
\caption{\label{tab:fine-graine} Results of fine-grained Attributes control on the Yelp. Different rows correspond to the set of attributes being controlled by the model.} 
\end{table*}

\subsection{Text Gender Style Transfer}
We use the same dataset\footnote{This dataset can be download at \url{http://tts.speech.cs.cmu.edu/style_models/gender_classifier.tar}.} as in \cite{prabhumoye2018style}, which contains reviews from Yelp annotated with two sexes (they only consider male or female due to the absence of corpora with other gender annotations). Following \cite{prabhumoye2018style}, we use the same pre-processing steps and similar experimental configurations. We directly compare our method against BST \cite{prabhumoye2018style} which has been shown to outperform the previous approach \cite{shen2017style} on this task. We use the same metrics described in Section Text Sentiment Transfer except for the BLEU score because this dataset does not provide the human annotated sentences. The implementation of BST is based on their source code\footnote{\url{https://github.com/shrimai/Style-Transfer-Through-Back-Translation}}. The results are shown in Table \ref{tab:gender}. We can see that our methods outperform BST \cite{prabhumoye2018style} on all metrics. The generated examples are shown in \textbf{Appendix C}.

\subsection{Multiple Fine-Grained Attributes Control}
To verify our method can also achieve multiple fine-grained attributes control, we take the attributes length, keyword presence, and sentiment as the case study in this experiment.
We use the same dataset, Yelp, and the same metrics used in Section Text Sentiment Transfer. For the attribute of length, we design two tasks: 1) We hope that the target sentence can add some relevant content to the original sentence, and increase its length by twice (denoted as Length$\Uparrow$); 2) We hope that the target sentence can compress the content of the original sentence and reduce its length by half (denoted as Length$\Downarrow$). For evaluation, we measure the percentage of the length of the generated sentences to the length of the original sentences (denoted as Len\%). For the attribute of keyword presence, we hope that the target sentence can contain a pre-defined keyword and retain the content of the original sentence as much as possible (denoted as Keywords). In our experiments, we define a keyword as a noun that is semantically most relevant (computed by the cosine distance of pre-trained word embeddings) to the original sentence but do not appear in the original sentence. The percentage of the generated sentences contain the pre-defined keyword (denoted as Key\%) is reported.

The results are shown in Table \ref{tab:fine-graine}. For a single fine-grained attribute, it can be observed that Keywords achieves 92.3 Key\% score, Length$\Uparrow$ and Length$\Downarrow$ achieve 208.8 and 40.8 Len\% scores respectively. At the same time, the fluency and content retention scores are still high. These results demonstrate the proposed method can control such fine-grained attributes. When we further control the sentiment attribute, we can see that Sentiment + Keywords achieves 91.6\% accuracy, while the accuracy of Sentiment + Length$\Uparrow$ and Sentiment + Length$\Downarrow$ is 97.7\% and 95.1\% respectively. Meanwhile, their rest scores have not declined significantly. When simultaneously controlling all these attributes, Sentiment + Keywords + Length$\Uparrow$ achieves 93.0\% accuracy, 183.7 Len\% score, and 66.6 Key\% score, while Sentiment + Keywords + Length$\Downarrow$ achieves 87.6\% accuracy, 60.9 Len\% score, and 63.0 Key\% score. Since it is more difficult to reduce sentence length than to increase sentence length while controlling other attributes, the fluency of Sentiment + Keywords + Length$\Downarrow$ is worse than Sentiment + Keywords + Length$\Uparrow$. We show some generated examples in \textbf{Appendix D}. These results indicate that our proposed method can control multiple attributes simultaneously.

\section{Related Works}
We have witnessed an increasing interest in text style transfer under the setting of non-parallel data. Most such methods explicitly disentangle the content and the attribute. One line of research leverages the auto-encoder framework to encode the original sentence into an attribute-independent content representation with adversarial learning, which is then fed into the decoder with a style vector to output the transferred sentence. 
In \cite{hu2017toward,shen2017style,prabhumoye2018style}, adversarial learning is utilized to ensure that the output sentence has the desired style. In order to disentangle the content and the attribute, \cite{hu2017toward} enforces the output sentence to reconstruct the content representation, while \cite{fu2018style,zhao2017learning,john2019disentangled} apply adversarial learning to discourage encoding style information into the content representation. \cite{shen2017style} utilizes adversarial learning to align the generated sentences from one style to the data domain of the other style. In \cite{yang2018unsupervised}, the authors extend the cross-align method \cite{shen2017style} by employing a language model as the discriminator, which can provide a more stable and more informative training signal for adversarial learning. 

However, as argued in \cite{li2018delete,lample2019rewriting}, it is often easy to fool the discriminator without actually learning the representations that are disentangled. Unlike the methods mentioned above that disentangle the content and the attribute with adversarial learning, another line of research \cite{prabhumoye2018style,logeswaran2018content,lample2019rewriting} applies back-translation \cite{ella16preserv} to rephrase a sentence while reducing the stylistic properties and encourage content compatibility. Besides, the authors in \cite{li2018delete} directly mask out the words associated with the original style of the sentence to obtain the attribute-independent content text.
Instead of revising the sentence in the discrete space with prior knowledge as in \cite{li2018delete}, our method maps the discrete sentence into a continuous representation space and revises the continuous representation with the gradient provided by the predictors. This method does not explicitly disentangle the content and the attribute and avoids the training difficulties caused by the use of adversarial learning in the previous methods. Similar ideas have been proposed in \cite{mueller2017sequence,luo2018neural} for sentence revision and neural architecture search. As pointed out in \cite{shen2017style}, the model proposed in \cite{mueller2017sequence} does not necessarily enforce content preservation, while our method employs a content predictor to enhance content preservation. Furthermore, unlike most previous methods that only control a single binary attribute (e.g., positive and negative sentiments), our approach can further control multiple fine-grained attributes such as sentence length and the existence of specific words. 
Note that controlling such fine-grained attributes has already been studied in the previous works for other tasks~\cite{post2018fast,makino2019global}, which only serves as a case study to demonstrate the generality of our method.

\section{Conclusion and Future Work}
In this paper, we propose a new framework for unsupervised text style transfer which revises the original sentences in a continuous space based on gradient optimization in the inference stage.
Compared with previous adversarial learning based methods, our method is easy to train, interpretable, and more controllable.
Extensive experiments on three popular text style transfer tasks show that our approach outperforms five previous state-of-the-art methods.
Furthermore, experimental results demonstrate that the proposed method can simultaneously manipulate multiple fine-grained attributes such as sentence length and the presence of specific words. 

In future work, we plan to explore control over other fine-grained attributes. In addition, it would be interesting to extend the proposed approach to other natural language generation tasks, such as dialogue and headline generation.

\section*{Acknowledgment}
This work is supported by National Key R\&D Program of China under contract No. 2017YFB1002201 and supported by National Natural Science Fund for Distinguished Young Scholar (Grant No. 61625204) and partially supported by the Key Program of National Science Foundation of China (Grant No. 61836006). 

\bibliography{4678.bibfile}

\begin{thebibliography}{}

\bibitem[\protect\citeauthoryear{Alemi \bgroup et al\mbox.\egroup
  }{2018}]{alemi2018fixing}
Alemi, A.; Poole, B.; Fischer, I.; Dillon, J.; Saurous, R.~A.; and Murphy, K.
\newblock 2018.
\newblock Fixing a broken elbo.
\newblock In {\em ICML}.

\bibitem[\protect\citeauthoryear{Bowman \bgroup et al\mbox.\egroup
  }{2016}]{bowman2015generating}
Bowman, S.~R.; Vilnis, L.; Vinyals, O.; Dai, A.~M.; Jozefowicz, R.; and Bengio,
  S.
\newblock 2016.
\newblock Generating sentences from a continuous space.
\newblock In {\em CoNLL}.

\bibitem[\protect\citeauthoryear{Fu \bgroup et al\mbox.\egroup
  }{2018}]{fu2018style}
Fu, Z.; Tan, X.; Peng, N.; Zhao, D.; and Yan, R.
\newblock 2018.
\newblock Style transfer in text: Exploration and evaluation.
\newblock In {\em AAAI}.

\bibitem[\protect\citeauthoryear{Goodfellow \bgroup et al\mbox.\egroup
  }{2014}]{Goodfellow2014Generative}
Goodfellow, I.; Pouget-Abadie, J.; Mirza, M.; Xu, B.; Warde-Farley, D.; Ozair,
  S.; Courville, A.; and Bengio, Y.
\newblock 2014.
\newblock Generative adversarial nets.
\newblock In {\em NIPS}.

\bibitem[\protect\citeauthoryear{Goyal \bgroup et al\mbox.\egroup
  }{2017}]{goyal2017z}
Goyal, A. G. A.~P.; Sordoni, A.; C{\^o}t{\'e}, M.-A.; Ke, N.~R.; and Bengio, Y.
\newblock 2017.
\newblock Z-forcing: Training stochastic recurrent networks.
\newblock In {\em NIPS}.

\bibitem[\protect\citeauthoryear{Guillaume~Lample}{2019}]{lample2019rewriting}
Guillaume~Lample, Sandeep~Subramanian, E. M. S. L. D. M. R. Y.-L.~B.
\newblock 2019.
\newblock Multiple attribute text rewriting.
\newblock In {\em ICLR}.

\bibitem[\protect\citeauthoryear{He and McAuley}{2016}]{he2016ups}
He, R., and McAuley, J.
\newblock 2016.
\newblock Ups and downs: Modeling the visual evolution of fashion trends with
  one-class collaborative filtering.
\newblock In {\em WWW}.

\bibitem[\protect\citeauthoryear{Hoffman and Johnson}{2016}]{hoffman2016elbo}
Hoffman, M.~D., and Johnson, M.~J.
\newblock 2016.
\newblock Elbo surgery: yet another way to carve up the variational evidence
  lower bound.
\newblock In {\em Workshop in Advances in Approximate Bayesian Inference,
  NIPS}.

\bibitem[\protect\citeauthoryear{Hu \bgroup et al\mbox.\egroup
  }{2017}]{hu2017toward}
Hu, Z.; Yang, Z.; Liang, X.; Salakhutdinov, R.; and Xing, E.~P.
\newblock 2017.
\newblock Toward controlled generation of text.
\newblock In {\em ICML}.

\bibitem[\protect\citeauthoryear{John \bgroup et al\mbox.\egroup
  }{2019}]{john2019disentangled}
John, V.; Mou, L.; Bahuleyan, H.; and Vechtomova, O.
\newblock 2019.
\newblock Disentangled representation learning for text style transfer.
\newblock In {\em AAAI}.

\bibitem[\protect\citeauthoryear{Kim}{2014}]{kim2014convolutional}
Kim, Y.
\newblock 2014.
\newblock Convolutional neural networks for sentence classification.
\newblock In {\em EMNLP}.

\bibitem[\protect\citeauthoryear{Kingma and Welling}{2013}]{kingma2013auto}
Kingma, D.~P., and Welling, M.
\newblock 2013.
\newblock Auto-encoding variational bayes.
\newblock In {\em ICLR}.

\bibitem[\protect\citeauthoryear{Kingma \bgroup et al\mbox.\egroup
  }{2016}]{kingma2016improved}
Kingma, D.~P.; Salimans, T.; Jozefowicz, R.; Chen, X.; Sutskever, I.; and
  Welling, M.
\newblock 2016.
\newblock Improved variational inference with inverse autoregressive flow.
\newblock In {\em NIPS}.

\bibitem[\protect\citeauthoryear{Kneser and Ney}{1995}]{kneser1995improved}
Kneser, R., and Ney, H.
\newblock 1995.
\newblock Improved backing-off for m-gram language modeling.
\newblock In {\em ICASSP}.

\bibitem[\protect\citeauthoryear{Kusner and
  Hern{\'a}ndez-Lobato}{2016}]{kusner2016gans}
Kusner, M.~J., and Hern{\'a}ndez-Lobato, J.~M.
\newblock 2016.
\newblock Gans for sequences of discrete elements with the gumbel-softmax
  distribution.
\newblock {\em arXiv preprint arXiv:1611.04051}.

\bibitem[\protect\citeauthoryear{Li \bgroup et al\mbox.\egroup
  }{2018}]{li2018delete}
Li, J.; Jia, R.; He, H.; and Liang, P.
\newblock 2018.
\newblock Delete, retrieve, generate: A simple approach to sentiment and style
  transfer.
\newblock In {\em NAACL-HLT}.

\bibitem[\protect\citeauthoryear{Logeswaran, Lee, and
  Bengio}{2018}]{logeswaran2018content}
Logeswaran, L.; Lee, H.; and Bengio, S.
\newblock 2018.
\newblock Content preserving text generation with attribute controls.
\newblock In {\em NeurIPS}.

\bibitem[\protect\citeauthoryear{Luo \bgroup et al\mbox.\egroup
  }{2018}]{luo2018neural}
Luo, R.; Tian, F.; Qin, T.; Chen, E.; and Liu, T.-Y.
\newblock 2018.
\newblock Neural architecture optimization.
\newblock In {\em NeurIPS}.

\bibitem[\protect\citeauthoryear{Makino \bgroup et al\mbox.\egroup
  }{2019}]{makino2019global}
Makino, T.; Iwakura, T.; Takamura, H.; and Okumura, M.
\newblock 2019.
\newblock Global optimization under length constraint for neural text
  summarization.
\newblock In {\em ACL}.

\bibitem[\protect\citeauthoryear{Melnyk \bgroup et al\mbox.\egroup
  }{2017}]{melnyk2017improved}
Melnyk, I.; Santos, C. N.~d.; Wadhawan, K.; Padhi, I.; and Kumar, A.
\newblock 2017.
\newblock Improved neural text attribute transfer with non-parallel data.
\newblock In {\em NIPS (Workshop)}.

\bibitem[\protect\citeauthoryear{Mueller, Gifford, and
  Jaakkola}{2017}]{mueller2017sequence}
Mueller, J.; Gifford, D.; and Jaakkola, T.
\newblock 2017.
\newblock Sequence to better sequence: continuous revision of combinatorial
  structures.
\newblock In {\em ICML}.

\bibitem[\protect\citeauthoryear{Och and Ney}{2004}]{och2004alignment}
Och, F.~J., and Ney, H.
\newblock 2004.
\newblock The alignment template approach to statistical machine translation.
\newblock {\em Computational linguistics}.

\bibitem[\protect\citeauthoryear{Papineni \bgroup et al\mbox.\egroup
  }{2002}]{papineni2002bleu}
Papineni, K.; Roukos, S.; Ward, T.; and Zhu, W.-J.
\newblock 2002.
\newblock Bleu: a method for automatic evaluation of machine translation.
\newblock In {\em ACL}.

\bibitem[\protect\citeauthoryear{Post and Vilar}{2018}]{post2018fast}
Post, M., and Vilar, D.
\newblock 2018.
\newblock Fast lexically constrained decoding with dynamic beam allocation for
  neural machine translation.
\newblock In {\em NAACL}.

\bibitem[\protect\citeauthoryear{Prabhumoye \bgroup et al\mbox.\egroup
  }{2018}]{prabhumoye2018style}
Prabhumoye, S.; Tsvetkov, Y.; Salakhutdinov, R.; and Black, A.~W.
\newblock 2018.
\newblock Style transfer through back-translation.
\newblock In {\em ACL}.

\bibitem[\protect\citeauthoryear{Roberts \bgroup et al\mbox.\egroup
  }{2018}]{RobertsERHE18}
Roberts, A.; Engel, J.; Raffel, C.; Hawthorne, C.; and Eck, D.
\newblock 2018.
\newblock A hierarchical latent vector model for learning long-term structure
  in music.
\newblock In {\em ICML}.

\bibitem[\protect\citeauthoryear{Schuster and
  Paliwal}{1997}]{schuster1997bidirectional}
Schuster, M., and Paliwal, K.~K.
\newblock 1997.
\newblock Bidirectional recurrent neural networks.
\newblock {\em IEEE Transactions on Signal Processing}.

\bibitem[\protect\citeauthoryear{Semeniuta, Severyn, and
  Barth}{2017}]{semeniuta2017hybrid}
Semeniuta, S.; Severyn, A.; and Barth, E.
\newblock 2017.
\newblock A hybrid convolutional variational autoencoder for text generation.
\newblock In {\em EMNLP}.

\bibitem[\protect\citeauthoryear{Shen \bgroup et al\mbox.\egroup
  }{2017}]{shen2017style}
Shen, T.; Lei, T.; Barzilay, R.; and Jaakkola, T.
\newblock 2017.
\newblock Style transfer from non-parallel text by cross-alignment.
\newblock In {\em NIPS}.

\bibitem[\protect\citeauthoryear{Shen \bgroup et al\mbox.\egroup
  }{2018}]{shen2018improving}
Shen, X.; Su, H.; Niu, S.; and Demberg, V.
\newblock 2018.
\newblock Improving variational encoder-decoders in dialogue generation.
\newblock In {\em AAAI}.

\bibitem[\protect\citeauthoryear{Sutton \bgroup et al\mbox.\egroup
  }{2000}]{Sutton1999Policy}
Sutton, R.~S.; McAllester, D.~A.; Singh, S.~P.; and Mansour, Y.
\newblock 2000.
\newblock Policy gradient methods for reinforcement learning with function
  approximation.
\newblock In {\em NIPS}.

\bibitem[\protect\citeauthoryear{Wintner \bgroup et al\mbox.\egroup
  }{2016}]{ella16preserv}
Wintner, S.; Mirkin, S.; Specia, L.; Rabinovich, E.; and Patel, R.~N.
\newblock 2016.
\newblock Personalized machine translation: Preserving original author traits.
\newblock In {\em EACL}.

\bibitem[\protect\citeauthoryear{Yang \bgroup et al\mbox.\egroup
  }{2017}]{yang2017improved}
Yang, Z.; Hu, Z.; Salakhutdinov, R.; and Berg-Kirkpatrick, T.
\newblock 2017.
\newblock Improved variational autoencoders for text modeling using dilated
  convolutions.
\newblock In {\em ICML}.

\bibitem[\protect\citeauthoryear{Yang \bgroup et al\mbox.\egroup
  }{2018}]{yang2018unsupervised}
Yang, Z.; Hu, Z.; Dyer, C.; Xing, E.~P.; and Berg-Kirkpatrick, T.
\newblock 2018.
\newblock Unsupervised text style transfer using language models as
  discriminators.
\newblock In {\em NeurIPS}.

\bibitem[\protect\citeauthoryear{Zhang \bgroup et al\mbox.\egroup
  }{2017}]{Zhang2017Adversarial}
Zhang, Y.; Gan, Z.; Fan, K.; Chen, Z.; Henao, R.; Shen, D.; and Carin, L.
\newblock 2017.
\newblock Adversarial feature matching for text generation.
\newblock In {\em ICML}.

\bibitem[\protect\citeauthoryear{Zhao, Zhao, and
  Eskenazi}{2017}]{zhao2017learning}
Zhao, T.; Zhao, R.; and Eskenazi, M.
\newblock 2017.
\newblock Learning discourse-level diversity for neural dialog models using
  conditional variational autoencoders.
\newblock In {\em ACL}.

\end{thebibliography}
\bibliographystyle{aaai}
\newpage
\section{A. Hyper-parameters and Ablation Study}
We study the effect of following hyper-parameters and configurations: 
\begin{enumerate}
  \item The hyper-parameter $\lambda_c$ described in Equation (12), which is the trade-off parameter to balance the content preservation and style transfer strength.
  \item The retraining of the predictors by Equation (6) and (10). We conduct an ablation study to verify the effect of the retraining.
  \item The target $x_\text{bow}$ of the content predictor. As described in Section 3.1, we proposed two kinds of $x_\text{bow}$: (a) Set the $x_\text{bow}$ to contain all the words in the original sentence $x$ (denoted as Cont-1); (b) Set the $x_\text{bow}$ to contain only all nouns (detected by a NLTK POS tagger) in $x$ (denoted as Cont-2). Besides, we test not using the content predictor (denoted as Cont-0).
  \item The KL loss of the Variational Auto-encoder (VAE). If the KL loss is too large, the VAE will collapse into an Auto-encoder. If the KL loss drops to 0, the VAE will collapse into a plain language model. Ideally, it should be small but non-zero \cite{hoffman2016elbo}. Under different configurations (e.g., KL annealing \cite{bowman2015generating}, and the weighted KL term \cite{alemi2018fixing}), we obtain VAEs of different KL losses and then test their performance in our scenarios.
\end{enumerate}
The rest of the hyper-parameters can be found in the source code \url{https://github.com/dayihengliu/Fine-Grained-Style-Transfer}. Table \ref{tab:study} reports these results on Yelp sentiment transfer task with the same settings in Section 4.1. From the results we can see:
\begin{enumerate}
  \item When the value of $\lambda_c$ is increased, the word overlap score and the Noun\% score increase while the sentiment transfer accuracy decreases. It demonstrates that the $\lambda_c$ can control the trade-off between attribute transfer and content preservation.
  \item After the retraining of sentiment predictor, the sentiment transfer accuracy increased from 88.3\% to 93.1\%. The retraining of content predictor further improves the word overlap score and the Noun\% score. These results show that the retraining of the predictors by Equation (6) and (10) can further improve the performance.
  \item As expected, Cont-1 can improve the word overlap score, while Cont-2 can further improve the Noun\% score and the sentiment transfer accuracy. Compared with the Cont-0, both Cont-1 and Cont-2 have significantly improved the success rate, which indicates the effectiveness of the content predictor.
  \item When the KL loss of the VAE is lower, the reconstruction error is higher. At the same time, the accuracy and the fluency are better, but the content preservation is poor. The KL term of VAE can also control the trade-off between attribute transfer and content preservation.
\end{enumerate}

\begin{table*}[h] \small
\begin{center}
\begin{tabular}{lcccccccccccc}
 \hline
Settings & Accuracy$\uparrow$ & PPL$\downarrow$ & Overlap$\uparrow$ & Noun\%$\uparrow$\\
\hline
1. $\lambda_c$: & \\
$\lambda_c$ = 0.1 & 88.7 & 20.6 & \textbf{35.7} & \textbf{73.6}\\ 
$\lambda_c$ = 0.05 & \textbf{93.4} & \textbf{19.8} & 34.8 & 68.6\\ 
\hline
2. Retraining: & \\
No Retraining & 88.3 & 19.4 & 39.4 & 58.6\\ 
+ Retrain Sentiment Predictor & 93.1 & 22.8 & 39.7 & 60.0\\ 
+ Retrain Content Predictor & \textbf{94.1} & \textbf{20.6} & \textbf{41.6} & \textbf{61.5} \\ 
 \hline
3. $x_\text{bow}$ type: & \\
Cont-0 & 91.9 & \textbf{12.6} & 33.2 & 43.4  \\ 
Cont-1 & 92.8 & 19.4 & \textbf{36.3} & 60.2 \\ 
Cont-2 & \textbf{93.4} & 19.8 & 35.7 & \textbf{68.6} \\ 
 \hline
4. KL loss: & \\
$\mathcal{L}_\text{KL}$ = 13.85 & \textbf{92.6} & \textbf{14.6} & 31.7 & 53.4 \\ 
$\mathcal{L}_\text{KL}$ = 17.27 & 88.8 & 19.8 & 38.1 & 69.1\\ 
$\mathcal{L}_\text{KL}$ = 21.84 & 84.7 & 27.6 & \textbf{43.1} & \textbf{86.8}\\ 
\hline
 \end{tabular}
\end{center}
\caption{\label{tab:study} Evaluation results of different hyper-parameters and configurations of the sentiment transfer task on Yelp. The notation $\uparrow$ means the higher is the better, while $\downarrow$ means the lower is the better.}
\end{table*}

\section{B. Samples of Sentiment Transfer} 
Some samples of the sentiment transfer task from ours and baselines on Yelp and Amazon are shown in Table \ref{tab:sentiment_yelp} and Table \ref{tab:sentiment_amazon}, respectively.
\begin{table*} \small
\begin{center}
\begin{tabular}{@{~~~~~}l@{~~~~~}l@{~~~~~}}
\toprule
\multicolumn{2}{c}{Sentiment transfer from \textbf{negative} to \textbf{positive} (Yelp)}\\
\toprule
Original &  we sit down and we got some really slow and lazy service . \\
Human & the service was quick and responsive . \\
CrossAligned & we went down and we were a good , friendly food . \\
MultiDecoder & we sit down and we got some really and fast food . \\
DeleteAndRetrieve & we got very nice place to sit down and we got some service . \\
BackTranslation & we got and i and it is very nice and friendly staff . \\
Ours (content-strengthen) & we sat down and got some really good service and friendly people . \\
Ours (balanced) & we sat down the street and had some really nice and fast service . \\
Ours (style-strengthen) & we really sit down and the service and food were great . \\
\hline
Original &  there was only meat and bread . \\
Human & there was a wide variety of meats and breads . \\
CrossAligned & there was amazing flavorful and . \\
MultiDecoder & there was only meat and bread . \\
DeleteAndRetrieve & meat and bread was very fresh . \\
BackTranslation & it was very nice and helpful . \\
Ours (content-strengthen) & the bread was fresh and the meat was tender . \\
Ours (balanced) & the bread was good and the bread was fresh and plentiful . \\
Ours (style-strengthen) & the bread was fresh and very tasty . \\
\hline
Original &  anyway , we got our coffee and will not return to this location . \\
Human & we got coffee and we 'll think about going back . \\
CrossAligned & anyway , we got our food and will definitely return to this location . \\
MultiDecoder & anyway , we got our coffee and will not return to this location . \\
DeleteAndRetrieve & anyway , we got our coffee and would recommend it to everyone . \\
BackTranslation & everything in the staff is very nice and it was the best . \\
Ours (content-strengthen) & i will return to this location , and we will definitely return . \\
Ours (balanced) & we will return to this location again , and the coffee was great . \\
Ours (style-strengthen) & we will definitely return , and this is our new favorite coffee place . \\
\toprule
\multicolumn{2}{c}{Sentiment transfer from \textbf{positive} to \textbf{negative} (Yelp)}\\
\toprule
Original &  i love this place , the service is always great ! \\
Human & hate this place , service was bad . \\
CrossAligned & i know this place , the food is just a horrible ! \\
MultiDecoder & i love this place , the service is always great ! \\
DeleteAndRetrieve & i did not like the homework of lasagna , not like it , . \\
BackTranslation & i wish i have been back , this place is a empty ! \\
Ours (content-strengthen) & however , this place is the worst i have ever been to . \\
Ours (balanced) & i do n't know why i love this place , but the service is horrible . \\
Ours (style-strengthen) & i do n't know why this place has the worst customer service ever . \\
\hline
Original &  their pizza is the best i have ever had as well as their ranch ! \\
Human & their pizza is the worst i have ever had as well as their ranch ! \\
CrossAligned & their pizza is the other i have ever had as well as their onions ! \\
MultiDecoder & their pizza is the best i have ever had as well at their job ! \\
DeleteAndRetrieve & had their bad taste like ranch ! \\
BackTranslation & their food is n't the worst i 've ever had to go ! \\
Ours (content-strengthen) & this is the worst pizza i have ever had as well as their ranch . \\
Ours (balanced) & this is the worst pizza i have ever had as well as their bruchetta . \\
Ours (style-strengthen) & i have had the worst pizza i have ever had in my life as well . \\
\hline
Original &  i will be going back and enjoying this great place ! \\
Human & i wo n't be going back and suffering at this terrible place ! \\
CrossAligned & i will be going back because from the \_num\_ stars place ! \\
MultiDecoder & i will be going back and often at no place ! \\
DeleteAndRetrieve & i will be going back and will not be returning into this anymore . \\
BackTranslation & i will not be going back and this place is awful ! \\
Ours (content-strengthen) & i will not be going back to this place for a while . \\
Ours (balanced) & i will not be going back to this place for a while . \\
Ours (style-strengthen) & i wo n't be going back to this place unless i 'm desperate . \\
\toprule
\end{tabular}
\end{center}
\caption{Samples of the sentiment transfer task from ours and baselines on Yelp. The Original denotes the input sentence, and the Human denotes the human annotated sentence. The samples of the sentiment transfer from negative to positive and positive to negative are shown in top and bottom, respectively.}\label{tab:sentiment_yelp}
\end{table*}

\begin{table*} \small
\begin{center}
\begin{tabular}{@{~~~~~}l@{~~~~~}l@{~~~~~}}
\toprule
\multicolumn{2}{c}{Sentiment transfer from \textbf{negative} to \textbf{positive} (Amazon)}\\
\toprule
Original &  ridiculous ! i had trouble getting it on with zero bubbles . \\
Human & great ! i had no trouble getting it on with zero bubbles . \\
CrossAligned & so far i have been using it for years and now . \\
MultiDecoder & beautiful i have to replace it with after using the \_num\_ \\
DeleteAndRetrieve & they are easy to use ,  i had trouble getting it on with zero bubbles . \\
BackTranslation & flavorful ! i don t have used it to work with \_num\_ years . \\
Ours (content-strengthen) & i have no trouble putting bubbles on it . \\
Ours (balanced) & i have had no trouble getting bubbles on it . \\
Ours (style-strengthen) & i ve had no problems with bubbles on it . \\
\hline
Original &  i ve used it twice and it has stopped working . \\
Human & used it without problems . \\
CrossAligned & i have it s so it s just work well . \\
MultiDecoder & i ve used it twice and it has gave together . \\
DeleteAndRetrieve & i ve used it twice and it has performed well . \\
BackTranslation & i ve been using this for \_num\_ years now and it works great . \\
Ours (content-strengthen) & i ve used it several times and it works great . \\
Ours (balanced) & i ve used it several times and it has worked flawlessly . \\
Ours (style-strengthen) & i ve used it for several months now and it has been working great . \\
\hline
Original &  i ve used these a few times and broke them very easily . \\
Human & i ve used these a few times and loved them . \\
CrossAligned & i ve had this for a few months and it s fine . \\
MultiDecoder & i ve used these a few times and use the iphone very quickly . \\
DeleteAndRetrieve & i ve used these a few times and broke them very easily !  . \\
BackTranslation & i ve had this case for \_num\_ years and it works great . \\
Ours (content-strengthen) & i ve used them a few times and they are very sturdy . \\
Ours (balanced) & i ve used them several times a week and they are very sturdy . \\
Ours (style-strengthen) & i ve used these a few times and they are very sturdy . \\
\toprule
\multicolumn{2}{c}{Sentiment transfer from \textbf{positive} to \textbf{negative} (Amazon)}\\
\toprule
Original &  this product does what it is suppose to do . \\
Human & this product does not do what it is supposed to do . \\
CrossAligned & this product isn t work and i have used . \\
MultiDecoder & this product does what it is supposed to do . \\
DeleteAndRetrieve & this product did not do what it was suppose to do . \\
BackTranslation & this product metropolis what it s like . \\
Ours (content-strengthen) & this product does not do what it claims to do . \\
Ours (balanced) & this product does not do what it claims to do . \\
Ours (style-strengthen) & this product does not do what it claims to do . \\
\hline
Original &  i would recommend to anyone who wants a pda . \\
Human & i would not recommend this to anyone who wants a pda . \\
CrossAligned & i would not recommend it to be a refund . \\
MultiDecoder & i would recommend to anyone who has it into . \\
DeleteAndRetrieve & i would not recommend this to anyone who wants a sensitive pda . \\
BackTranslation & i wish i would give them a lot of them . \\
Ours (content-strengthen) & i would not recommend this product to anyone . \\
Ours (balanced) & i would not recommend this to anyone who wants a <UNK> . \\
Ours (style-strengthen) & i would not recommend this to anyone who wants a <UNK> . \\
\hline
Original &  i have been extremely happy with my purchase . \\
Human & upset at purchase from the start . \\
CrossAligned & i have been using them for my hair . \\
MultiDecoder & i have been extremely happy with my review . \\
DeleteAndRetrieve & i have been extremely disappointed with this purchase . \\
BackTranslation & i was very disappointed with my phone . \\
Ours (content-strengthen) & i am very disappointed with this purchase and would not purchase again . \\
Ours (balanced) & i have been extremely disappointed with my purchase . \\
Ours (style-strengthen) & i am very disappointed with this purchase .  \\
\hline
\toprule
\end{tabular}
\end{center}
\caption{Samples of the sentiment transfer task from ours and baselines on Amazon. The Original denotes the input sentence, and the Human denotes the human annotated sentence. The samples of the sentiment transfer from negative to positive and positive to negative are shown in top and bottom, respectively.}\label{tab:sentiment_amazon}
\end{table*}

\section{C. Samples of Gender Style Transfer}
Table \ref{tab:gender_yelp} shows some samples of the gender style transfer task from ours and the strong baseline. 

\begin{table*} \small
\begin{center}
\begin{tabular}{@{~~~~~}l@{~~~~~}l@{~~~~~}}
\toprule
\multicolumn{2}{c}{Gender style transfer from \textbf{male} to \textbf{female}}\\
\toprule
Original &  i wish there is more than 0 stats to give you . \\
BackTranslation &  i think there ' s than 0 stars to see you . \\
Ours (content-strengthen) &  i wish i could give more stars . \\
Ours (balanced) &  i wish there would give more stars . \\
Ours (style-strengthen) &  i wish i could give more stars . \\
\hline
Original &  good vibe , good drinks and prices and unique decoration . \\
BackTranslation &  good service , good service , and the service and décoration . \\
Ours (content-strengthen) &  overall , the drinks were really good . \\
Ours (balanced) &  overall , the drinks are really good and unique . \\
Ours (style-strengthen) &  the drinks are good , and the decor is cute . \\
\hline
Original &  the food was n't anything outstanding to justify the price . \\
BackTranslation &  the food was kind of a good time to try the price . \\
Ours (content-strengthen) &  i ca n't wait to go back and the food was n't anything special . \\
Ours (balanced) &  the food was n't anything special . \\
Ours (style-strengthen) &  the food was n't anything special . \\
\hline
Original &  the cost was more for the size than the quality . \\
BackTranslation &  the service itself was very good for the price that ' s hotels . \\
Ours (content-strengthen) &  the portion size was more than enough for me . \\
Ours (balanced) &  the portion size was more than enough for the size . \\
Ours (style-strengthen) &  the portion size was more than \$ \_num\_ for the size of the portion . \\
\toprule
\multicolumn{2}{c}{Gender style transfer from \textbf{female} to \textbf{male}}\\
\toprule
Original &  we went here for my fiance ' s birthday . \\
BackTranslation &  we went here for my wife ' s anniversaire . \\
Ours (content-strengthen) &  went here for my wife ' s birthday . \\
Ours (balanced) &  went here for my wife ' s birthday . \\
Ours (style-strengthen) &  went here for my wife ' s birthday . \\
\hline
Original &  they always take such good care of me . \\
BackTranslation &  they always do a good job . \\
Ours (content-strengthen) &  they do a good job of taking care of you . \\
Ours (balanced) &  they always take care of you . \\
Ours (style-strengthen) &  they do a good job of taking care of you . \\
\hline
Original &  if you do come for breakfast get a croissant . \\
BackTranslation &  if you are looking for lunch , has a stems . \\
Ours (content-strengthen) &  if you come here for breakfast , you get a breakfast sandwich . \\
Ours (balanced) &  do n't come here if you want a breakfast sandwich . \\
Ours (style-strengthen) &  breakfast croissant is a must if you come here for breakfast . \\
\hline
Original &  the only thing worth mentioning was their dessert . \\
BackTranslation &  only compared to say was their service . \\
Ours (content-strengthen) &  the only thing worth mentioning is the deserts . \\
Ours (balanced) &  the only thing worth mentioning is the deserts . \\
Ours (style-strengthen) &  the only thing worth mentioning is the dessert . \\
\toprule
\end{tabular}
\end{center}
\caption{Samples of the gender style transfer task from ours and baselines. The Original denotes the input sentence. The samples of the gender style transfer from male to female and female to male are shown in top and bottom, respectively.}\label{tab:gender_yelp}
\end{table*}
\section{D. Samples of Multiple Fine-Grained Attributes Control}
The samples of multiple fine-grained attributes control are shown in Table \ref{tab:multi}.

\begin{table*} \scriptsize
\begin{center}
\begin{tabular}{@{~~~~~}l@{~~~~~}l@{~~~~~}}
\toprule
\multicolumn{2}{c}{Multiple fine-grained attributes control (from \textbf{negative} to \textbf{positive})}\\
\toprule
Original  &  i was very disappointed with this place . \\
Keywords  &  i \textbf{love} this place . \\
Sentiment + Keywords  &  i \textbf{love} this place too . \\
Length$\Uparrow$  &  i \textbf{love} this place , and i 'm so glad i went to the house . \\
Sentiment + Length$\Uparrow$  &  i was very disappointed with this place , and i was not impressed with it . \\
Keywords + Length$\Uparrow$  &  i was very impressed with this place and this place was very good . \\
Sentiment + Keywords + Length$\Uparrow$  &  i was very impressed with . \\
Length$\Downarrow$  &  very disappointed overall . \\
Sentiment + Length$\Downarrow$  &  \textbf{love} this . \\
Keywords + Length$\Downarrow$  &  i was very impressed with this place and \textbf{love} this place . \\
Sentiment + Keywords + Length$\Downarrow$  &  i \textbf{love} this place . \\
\hline
Original  &  at this location the service was terrible . \\
Keywords  &  the location at location was very \textbf{convenient} . \\
Sentiment + Keywords  &  the location is \textbf{convenient} and \textbf{convenient} . \\
Length$\Uparrow$  &  the location at this location was \textbf{convenient} and the service was horrible . \\
Sentiment + Length$\Uparrow$  &  this was the first time i went to this location and the service was terrible . \\
Keywords + Length$\Uparrow$  &  the service at this location was great and the food was very good . \\
Sentiment + Keywords + Length$\Uparrow$  &  the service at this location . \\
Length$\Downarrow$  &  terrible customer service . \\
Sentiment + Length$\Downarrow$  &  this location is \textbf{convenient} . \\
Keywords + Length$\Downarrow$  &  the location at this location is great and the location is very \textbf{convenient} . \\
Sentiment + Keywords + Length$\Downarrow$  &  location is \textbf{convenient} . \\
\hline
Original  &  i 'll keep looking for a different salon . \\
Keywords  &  i \textbf{love} looking for this nail salon . \\
Sentiment + Keywords  &  i \textbf{love} this nail salon for sure . \\
Length$\Uparrow$  &  i \textbf{love} this place , and i 'll be looking for a new nail . \\
Sentiment + Length$\Uparrow$  &  i 'll be looking for a different nail salon , and i do n't know . \\
Keywords + Length$\Uparrow$  &  i have been to this salon for a couple of the day , and it 's always the same thing i have ever made . \\
Sentiment + Keywords + Length$\Uparrow$  &  \textbf{love} this salon . \\
Length$\Downarrow$  &  definitely a salon . \\
Sentiment + Length$\Downarrow$  &  i \textbf{love} this nail salon . \\
Keywords + Length$\Downarrow$  &  i \textbf{love} this place , and i 'll be looking for a new nail . \\
Sentiment + Keywords + Length$\Downarrow$  &  i \textbf{love} this nail salon . \\
\toprule
\multicolumn{2}{c}{Multiple fine-grained attributes control (from \textbf{positive} to \textbf{negative})}\\
\toprule
Original  &  the best mexican food in the phoenix area . \\
Keywords  &  this is the best mexican \textbf{restaurant} in the area . \\
Sentiment + Keywords  &  this was the worst chinese \textbf{restaurant} in the phoenix area . \\
Length$\Uparrow$  &  this is the best mexican food i have had in the area and the area . \\
Sentiment + Length$\Uparrow$  &  this was the worst chinese food i have had in the phoenix area in phoenix . \\
Keywords + Length$\Uparrow$  &  this is the best mexican food in the area and the best \textbf{restaurant} in phoenix . \\
Sentiment + Keywords + Length$\Uparrow$  &  this was the worst chinese \textbf{restaurant} i have ever been to in the entire area . \\
Length$\Downarrow$  &  best mexican food . \\
Sentiment + Length$\Downarrow$  &  the worst food in phoenix . \\
Keywords + Length$\Downarrow$  &  best mexican \textbf{restaurant} in phoenix . \\
Sentiment + Keywords + Length$\Downarrow$  &  the worst \textbf{restaurant} in the area . \\
\hline
Original  &  thank you amanda , i will be back ! \\
Keywords  &  \textbf{thanks} again , thank you angela ! \\
Sentiment + Keywords  &  no \textbf{thanks} , i will not be back . \\
Length$\Uparrow$  &  if you are in the mood , i will definitely be taking care of you . \\
Sentiment + Length$\Uparrow$  &  if you want to be treated rudely , i will be taking care of you . \\
Keywords + Length$\Uparrow$  &  \textbf{thanks} to steven , i will be back , thank you for my next experience ! \\
Sentiment + Keywords + Length$\Uparrow$  &  if i asked him , i will be taking my car elsewhere , no \textbf{thanks} . \\
Length$\Downarrow$  &  thank you ! \\
Sentiment + Length$\Downarrow$  &  i will not be back . \\
Keywords + Length$\Downarrow$  &  \textbf{thanks} again , thank you ! \\
Sentiment + Keywords + Length$\Downarrow$  &  no \textbf{thanks} , thank you ! \\
\hline
Original  &  service was great and food was even better . \\
Keywords  &  terrible \textbf{customer} service and even better customer service . \\
Sentiment + Keywords  &  the \textbf{customer} service was terrible even worse than it was . \\
Length$\Uparrow$  &  the food was great , and the service was even better than i remembered it . \\
Sentiment + Length$\Uparrow$  &  the food was terrible and the service was even worse than it was even worse . \\
Keywords + Length$\Uparrow$  &  the \textbf{customer} service was terrible and the food was even worse than i remembered it . \\
Sentiment + Keywords + Length$\Uparrow$  &  the \textbf{customer} service was terrible even though it was n't even worse than before . \\
Length$\Downarrow$  &  service was great . \\
Sentiment + Length$\Downarrow$  &  the service was even worse . \\
Keywords + Length$\Downarrow$  &  \textbf{customer} service was even better . \\
Sentiment + Keywords + Length$\Downarrow$  &  even worse \textbf{customer} service was terrible . \\
\hline
\toprule
\end{tabular}
\end{center}
\caption{Samples of multiple fine-grained attributes control from ours. We bold the pre-defined keyword.}\label{tab:multi}
\end{table*}
\end{document}